# A data set for evaluating the performance of multi-class multi-object video tracking

Avishek Chakraborty[a], Victor Stamatescu[a], Sebastien C. Wong[b], Grant Wigley[a], David Kearney[a]

[a]Computational Learning Systems Laboratory, School of Information Technology and Mathematical Sciences, University of South Australia, Mawson Lakes, SA, Australia; [b]Defence Science and Technology Group, Edinburgh, SA, Australia

**ABSTRACT**

One of the challenges in evaluating multi-object video detection, tracking and classification systems is having publically available data sets with which to compare different systems. However, the measures of performance for tracking and classification are different. Data sets that are suitable for evaluating tracking systems may not be appropriate for classification. Tracking video data sets typically only have ground truth track IDs, while classification video data sets only have ground truth class-label IDs. The former identifies the same object over multiple frames, while the latter identifies the type of object in individual frames. This paper describes an advancement of the ground truth meta-data for the DARPA Neovision2 Tower data set to allow both the evaluation of tracking and classification. The ground truth data sets presented in this paper contain unique object IDs across 5 different classes of object (Car, Bus, Truck, Person, Cyclist) for 24 videos of 871 image frames each. In addition to the object IDs and class labels, the ground truth data also contains the original bounding box coordinates together with new bounding boxes in instances where un-annotated objects were present. The unique IDs are maintained during occlusions between multiple objects or when objects re-enter the field of view. This will provide: a solid foundation for evaluating the performance of multi-object tracking of different types of objects, a straightforward comparison of tracking system performance using the standard Multi Object Tracking (MOT) framework, and classification performance using the Neovision2 metrics. These data have been hosted publically.

## 1. INTRODUCTION

There are two fundamental tasks that a general purpose automated vision application may need to perform: (1) multi-object tracking (MOT) and (2) object recognition. MOT involves finding and following the same object over multiple frames of a video sequence and providing this object with a unique track identifier (ID). Object recognition involves automatically assigning the correct class label ID to the object. These are two different but related problems. If for a particular vision application only one type of object is of interest, then only the MOT task needs to be solved. An example of this is pedestrian tracking for visual surveillance where it is important to monitor the path of a suspicious individual through a crowded scene. If there is no requirement to associate the same object across multiple frames and only the class of object is of interest, then this becomes an object recognition task. An example of this is counting the numbers of each type of product on a warehouse shelf. However, there exist applications where it is important to solve both of these tasks simultaneously. For example, imagine a self-driving car application, where the vision system needs to maintain persistent tracks on objects, so trajectories can be estimated to avoid collisions, as well as recognise the class of object (street-sign, pedestrian, car) to understand its likely behaviour.

These two tasks can be treated separately. This separation has led to disparate research communities with specialised data sets for evaluating (1) multi-object tracking[1,2], and (2) object recognition[3]. This approach has led to measurable advancements of state-of-the-art for each of these two tasks[4,5].

It is also possible to integrate the two tasks to form a general purpose vision system, where multi-object tracking and object recognition components are interwoven[6,7]. However, for such a system, using separate benchmarks for the evaluation of tracking performance and recognition performance becomes impractical. For example, evaluating the tracking performance of such a system against the MOT Challenge[1,2] data sets, where pedestrians are the objects of interest, would not adequately assess the ability of the system to also track cars. Moreover, these would be considered false alarm tracks, unless they were first filtered out by the evaluation framework. Similarly using an image classification data set, such as ImageNet[3], would not exercise the ability of the system to locate multiple objects across multiple frames.

This motivates the need for a benchmark data set that considers both of these tasks holistically. This data set would be valuable in evaluating multi-class, multi-object detection, tracking and classification systems.

In this paper, we describe the development of a data set for evaluating multi-class, multi-object tracking algorithms. This is achieved by augmenting the existing Neovision2 Tower data set, which was created specifically for online object recognition[8]. The resulting data set enables the evaluation of both multi-object tracking and online recognition algorithms.

The rest of this paper is organised as follows. Section 2 describes the existing benchmarks that are used solely for performance evaluation of multi-object tracking or recognition. Section 3 describes in detail the Neovision2 Tower data set. Section 4 presents the approach taken to advance the existing the Neovision2 Tower data set to enable performance evaluation of multi-class multi-object tracking algorithms. Section 5 describes how the resulting augmented data set may be used in the evaluation of multi-object tracking algorithms. Finally, section 6 concludes this paper.

## 2. COMPARISON WITH EXISTING BENCHMARKS

This section outlines the motivation for choosing to work with the Neovision2 benchmark[9]. Existing large-scale data sets are either aimed at tracking[1,2,10] or recognition tasks[3]. In terms of video analysis, there exist three dedicated multi-object benchmarks: Neovision2[9], MOT Challenge[1,2] and UA-DETRAC[10]. These data sets are compared in Table 1 and in Figure 1.

Table 1. Public multi-object tracking and object recognition data sets. MOT Challenge and UA-DETRAC statistics are from[10].

| Benchmark | MOT Challenge | MOT Challenge | UA-DETRAC | Neovision2 | Neovision2 |
|---|---|---|---|---|---|
| Data set | MOT 2015 [1] | MOT16 [2] | UA-DETRAC [10] | Neovision2 Tower [9] | Neovision2 Heli [9] |
| Bounding-box format | Axis aligned | Axis aligned | Axis-aligned | Oriented | Oriented |
| Purpose of the data set | MOT | MOT | Detection & MOT | Detection & Recognition | Detection & Recognition |
| Number of videos | 11 train<br>11 test | 7 train<br>7 test | 60 train<br>40 test | 50 train<br>50 test | 32 train<br>37 test |
| Number of frames | 5.5k train<br>5.8k test | 5.3k train<br>5.9k test | 84k train<br>66k test | 43.6k train<br>43.6k test | 11k train<br>13k test |
| Number of classes | 1 | 1 (12 available) | 1 (4 available) | 5 | 10 |

The MOT 2015 and MOT16 data sets, which are intended for multi-object tracking, consider only one predominant class: pedestrians. In the case of MOT16, the meta-data contains 12 ground truth class labels that span three broad object categories: *Target* (pedestrian, person on vehicle), *Ambiguous* (static person, distractor, reflection) and *Other* (occluder, occluder full, occluder on the ground, car, bicycle, motorbike, non-motorized vehicle). The MOT16 evaluation framework, however, filters out all objects not classified as pedestrian.

The UA-DETRAC benchmark seeks to evaluate both detection and tracking of multiple objects. A critical limitation of the data set is that the vast majority of the target objects are moving. Hence, the detection and tracking of stationary objects are seldom evaluated. Furthermore, some stationary vehicles are labelled as 'ignore' while others are labelled as ground-truth objects. The ground truth meta-data contains four vehicle class labels: car, bus, van, others. Any pedestrians or cyclists are not annotated. In a similar approach to MOT16, the UA-DETRAC evaluation framework considers all vehicles as a single target class: vehicle.

The DARPA Neovision2 benchmark contains two public data sets, Tower and Heli, which were captured using a stationary camera and a camera mounted on a flying helicopter, respectively. Neovision2 Tower contains five target class labels (bus, truck, car, cyclist, person) annotated in every video frame, while Neovision2 Heli contains ten target class labels (car, truck, tractor-trailer, bus, container, boat, airplane, helicopter, person, cyclist) annotated in every fourth video frame. The Neovision2 Tower data set is a good candidate for evaluating an automated vision application because it contains different types of objects, both moving and stationary, annotated in every frame. A limitation of the Neovision2 benchmark, however, is that the meta-data does not include unique object identifiers. This allows for the evaluation of object detection and recognition, but not object tracking.

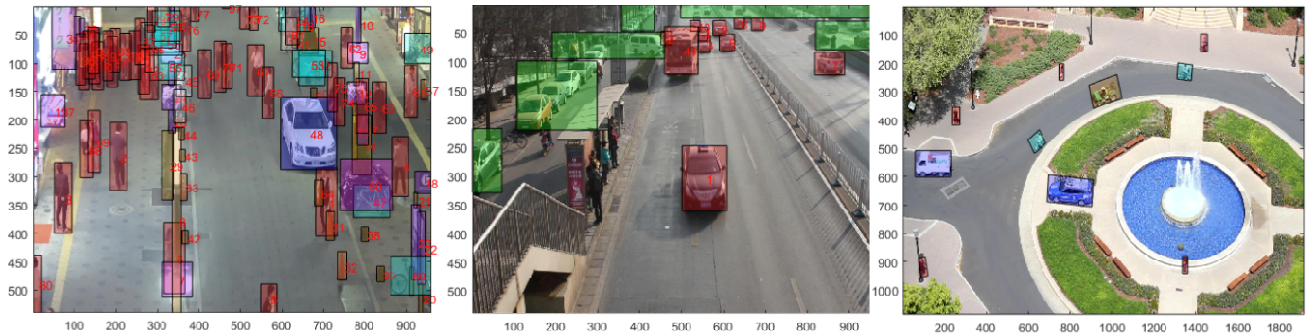

Figure 1. Illustration of the key differences between the MOT16 (left), UA-DETRAC (center) and Neovision2 benchmarks (right). MOT16 and UA-DETRAC both contain unique object IDs, with the meta-data of the latter also containing so-called ignore regions (shown in green), which are used to filter out detections in those parts of the scene. By contrast, Neovision2 was purpose built for online object recognition and does not possess unique object IDs.

## 3. NEOVISION2 TOWER DATA SET

The Neovision2 Tower data set comprises a set of videos that were captured from a fixed camera placed on the Stanford University Hoover tower, recorded at 29.97 frames/s with a resolution of 1920 x 1088. The images were converted to 8-bit PNG frames, resulting in 50 training and 50 test image sequences. Kasturi et al.[8] provide the public download link[9] to the training data set and information on obtaining the privately hosted test data set from the curators. Each video consists of 898 frames and a corresponding CSV file for the ground-truth meta-data. Each CSV file contains the ground-truth annotations and class labels for the first 871 of 898 frames, and we consider the benchmark as being limited to these annotated frames.

Human annotators were employed to draw the bounding boxes for every target object and identify their object class. This information was recorded in the ground-truth meta-data, together with additional information, namely: confidence of classification, degree of occlusion, ambiguity, site information, and version number. Table 2 describes each of these items in more detail.

While using this data set, it was found that video 04 contained a non-annotated target object throughout the entire image sequence. These missing annotations were corrected using a MATLAB based tool, which was purpose-built to interactively generate the missing bounding boxes. The tool allows recording of the coordinates for the mouse-clicked points, which are then stored in the appropriate row and column of the corresponding CSV file. This tool has been used to generate the coordinates of a bounding box that would surround the target object. The augmented ground-truth data, which is presented in the next section, is based on these corrected ground-truth annotations. An example of the difference before and after annotation is shown for video 04 in Figure 2.

Table 2. Description of the ground-truth meta-data for each object in the Neovision2 Tower data set.

| Item | Type | Description |
|---|---|---|
| **Frame** | Decimal | Frame number where the object is present |
| **4 Bounding Box Coordinates** | Decimal | There are 8 items for X and Y coordinates of the bounding box. |
| **ObjectType** | String | This may be: 'Bus', 'Truck', 'Car', 'Cyclist', 'Person' or an empty string if the object is 'Unknown' and cannot be classified |
| **Occlusion** | Boolean | If the object was occluded in this frame |
| **Ambiguous** | Boolean | Indicates if the object type is ambiguous |
| **Confidence** | Float | The confidence of the classification ranging from 0 to 1.0, where 1.0 shows complete certainty of the classification. |
| **SiteInfo** | | Indicates the site location |
| **Version** | Float | This is the version number, which can be empty |

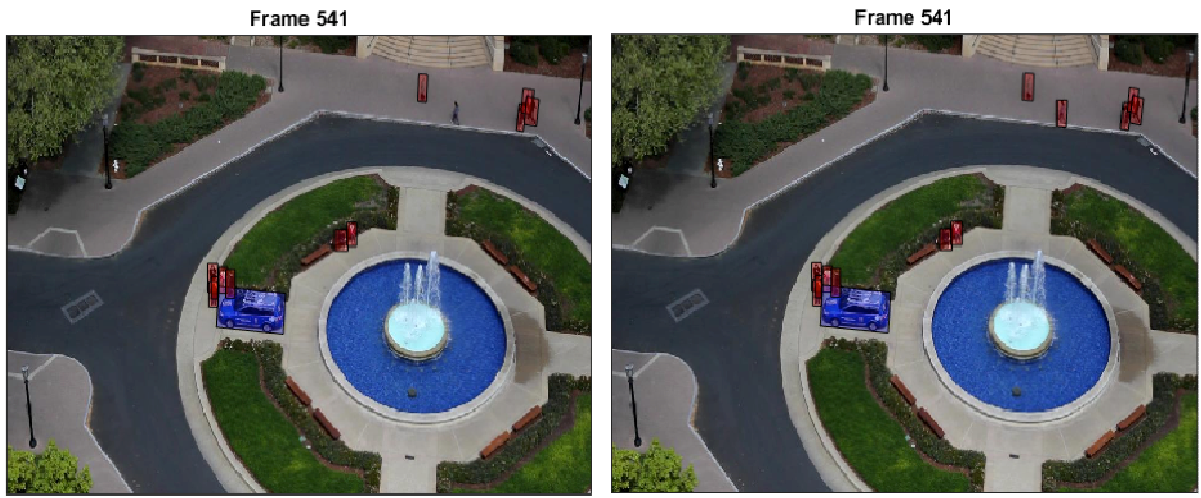

Figure 2. The image on the left shows the original data set where a pedestrian (near the top of the image) was not annotated with a bounding box. This has been rectified, as shown in the image on the right.

The Neovision2 Tower data set contains five object classes: Bus, Truck, Car, Cyclist, Person as well as an Unknown class for moving objects that do not fit into these categories and are to be filtered out by the evaluation tool. The abundances of these different object classes across all image sequences considered in this paper are shown in Figure 3. Some image sequences contain object occlusions by neighbouring objects or background clutter, as well as the re-appearance of target objects that had previously left the scene. These characteristics make the Neovision2 Tower data set a suitable candidate to evaluate multi-object tracking algorithms in a multi-class scenario. However, the ground-truth data required augmentation with unique object IDs to allow for the evaluation of tracking performance.

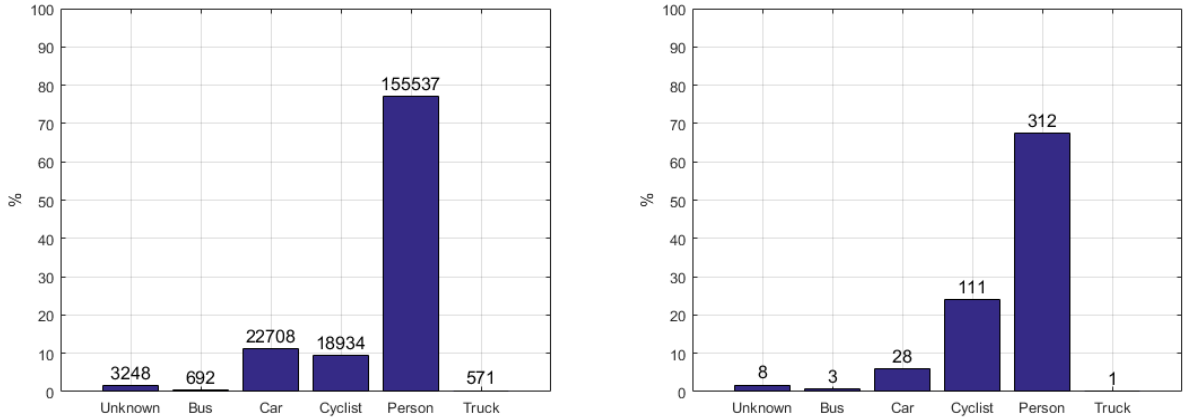

Figure 3. Object class abundances over Neovision2 Tower training videos 01-24, each comprising 871 annotated image frames, (left) considering all 201690 object annotations and, (right) considering all 463 unique objects (see Section 4).

The challenges for both tracking and recognition are related to the size of the objects of interest and to their occlusion by other objects or background clutter. In particular the degree of object occlusion can lead to reduced tracking and recognition performance. The maximum occlusion ratio of an object may be defined as the largest fraction of its bounding box being occluded by one of its neigbouring objects[10]. Typically, larger objects are less likely to be occluded than smaller objects. The size of an object is represented by its scale, which is calculated as the square root of the bounding box area, in units of pixels. Figures 4 and 5 illustrate the object scale and maximum occlusion ratio, respectively, for image sequences 01-24 of the Neovision2 Tower training data set. The most difficult class for both tracking and classification in this data set is likely to be Person, as it has both the smallest scale and highest maximum occlusion ratio of any class. Furthermore, we note that the distributions shown in Figure 5 do not consider occlusion from background objects, such as trees and lampposts.

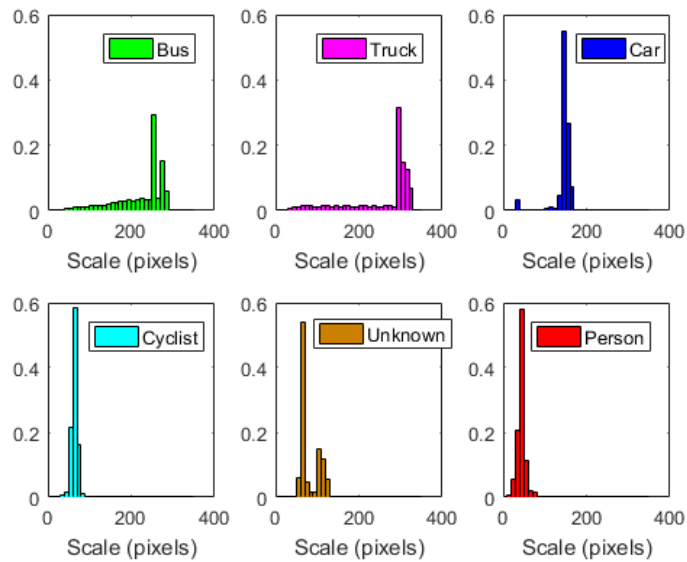

Figure 4. Normalized distributions of the scale (in units of pixels) occupied by objects of each class.

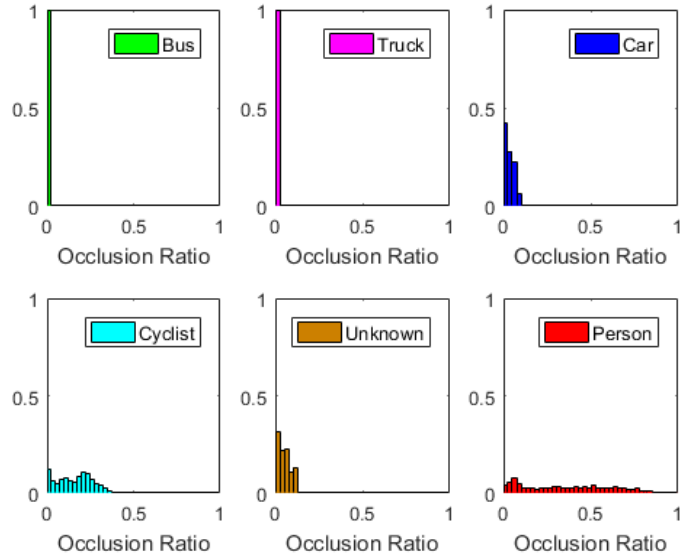

Figure 5. Normalized distributions of the maximum occlusion ratio of any object due to neighbouring objects in the same frame.

## 4. AUGMENTING THE GROUND-TRUTH DATA

In order to enable the quantitative evaluation of tracking performance, the ground-truth data needs to uniquely identify every target object and this uniqueness must be maintained across all the frames within the image sequence. This is achieved by augmenting the existing ground-truth data with an item to represent unique IDs for every target object. Table 3 provides an example of the resulting ground-truth for object ID 3, corresponding to the blue car in Figure 6.

Table 3. Example of ground-truth data for object ID 3 shown in Figure 6 and Frame 194

| Frame | 194 |
|---|---|
| **BoundingBox_X1** | 693 |
| **BoundingBox_Y1** | 594 |
| **BoundingBox_X2** | 922 |
| **BoundingBox_Y2** | 608 |
| **BoundingBox_X3** | 916 |
| **BoundingBox_Y3** | 708 |
| **BoundingBox_X4** | 687 |
| **BoundingBox_Y4** | 694 |
| **ObjectType** | Car |
| **Occlusion** | FALSE |
| **Ambiguous** | FALSE |
| **Confidence** | 1 |
| **SiteInfo** | |
| **Version** | 1.4 |
| **ID** | 3 |

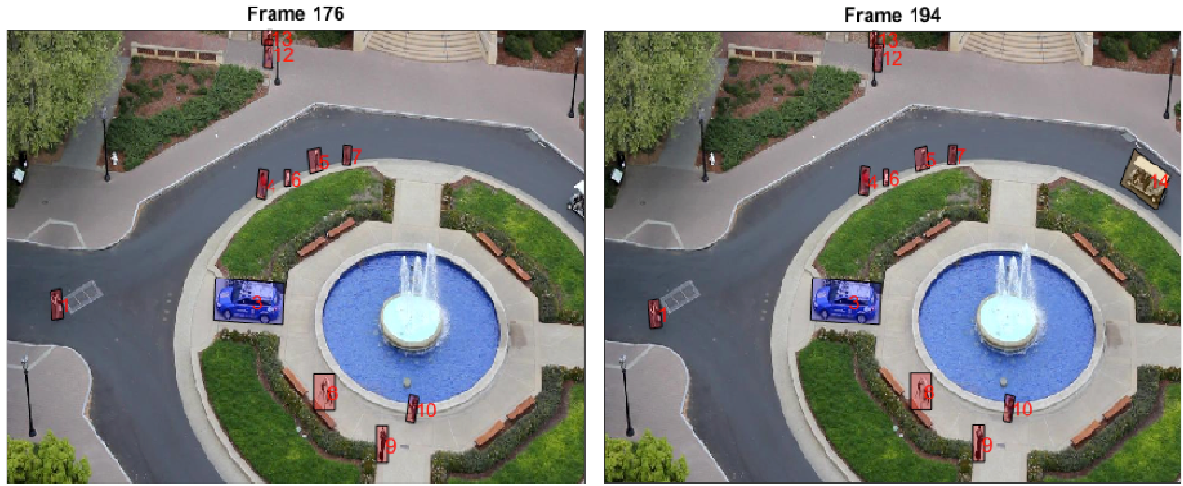

Figure 6. Frame 176 shows a new vehicle, which is unclassified entering the scene. After it has appeared, the number '14' is assigned as its unique ID

The policies undertaken to create these unique IDs are as follows. The unique IDs are sequential integers starting with 1 and as new objects appear within a certain video, they are incremented. In the first frame, assume that there are N objects. Then every object will be assigned an ID that will range from 1 to N. This initial assignment of an ID with an object is arbitrary. Now in the second frame if two new objects enter the scene, then they will be assigned N+1 and N+2 as their IDs. This procedure is repeated for each of the 871 annotated frames for every selected video. It must be noted that to maintain uniqueness, IDs are not recycled when an older object leaves the scene. Given that the ID creation was conducted manually, it was possible to identify objects leaving and then re-entering the scene. Once an object re-enters, the former ID is re-assigned. A dedicated MATLAB script was used to visualise the IDs for every object. This is illustrated in Figure 6, which shows two frames from the same video that demonstrate how an ID is assigned to a new target object. Two individuals verified the uniqueness of the IDs independently by visualizing every sequence.

The unique IDs also allow to compute the average speed of every unique object across its image sequence. The class-wise distributions of average speed in Figure 7 show that the data set includes stationary pedestrians and a stationary car, which present a challenge for systems whose object detection relies solely on motion-based image features.

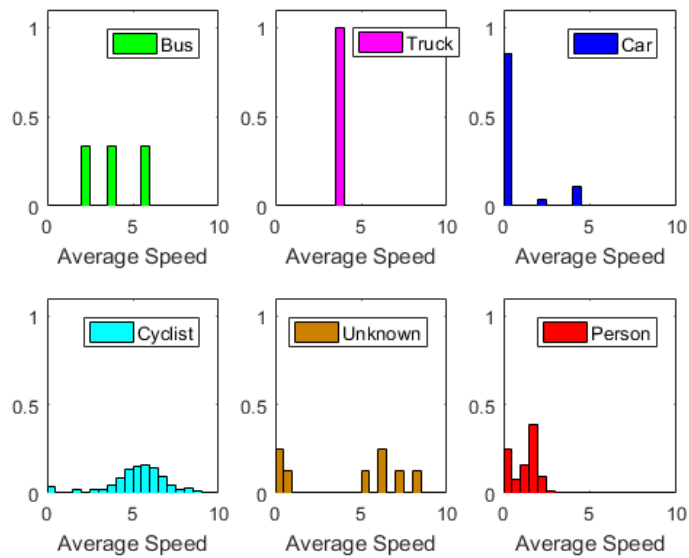

Figure 7. Normalized distributions of the average speed (in units of pixels/frame) of unique objects.

Although the Neovision2 Tower data set consists of 50 videos, we limit ourselves to the first 24 videos due to resource constraints associated with human annotation. This resulted in 20904 annotated image frames, containing 201690 object annotations and 463 unique objects. When compared with the MOT Challenge data sets[1,2], this presents a sizeable data set for training and evaluation of multi-class multi-object algorithms.

## 5. USAGE OF THE AUGMENTED GROUND TRUTHS

This section describes how the new track IDs in Neovision2 Tower may be used to evaluate multi-class, multi-object tracking systems. To date, the Neovision2 detection metrics[8] have been applied to the original Neovision2 Tower data set to evaluate object detection systems by considering all objects as one target class, and object recognition systems, which evaluates the detection of each object class separately.

Using our new track IDs together with the CLEAR MOT metrics[11], it now becomes possible to evaluate the performance of multi-object tracking systems across all object types and/or to evaluate the performance for each object class separately. The latter requires filtering out all ground truth and system tracks that correspond to all but the class under consideration.

Both the Neovision2 and CLEAR MOT metrics are based on the degree of spatial overlap $d_t$ between each ground truth bounding box region $R_{i,t}$ and each candidate bounding box region $R_{j,t}$ given by the $j^{th}$ system track at frame $t$:

$$d_t = (R_{i,t} \cap R_{j,t}) / (R_{j,t} \cup R_{j,t}) \quad (1)$$

For data sets whose ground truth meta-data includes unique object IDs, multi-class multi-object tracking performance can be evaluated by applying the CLEAR MOT metrics implementation provided by MOT Challenge[12]. This uses the Munkres algorithm[13] to find an optimum mapping (in terms of total spatial overlap) between system tracks and unique ground truth objects. The uniquely associated ($R_{i,t}$, $R_{j,t}$) pairs are identified as matches when $d_t$ exceeds a user-defined threshold $T_d$, which can range between 0 and 1. The CLEAR MOT formalism provides two complementary measures of system performance: Multiple Object Tracking Precision (*MOTP*) and Multiple Object Tracking Accuracy (*MOTA*). *MOTP* is calculated by averaging the matched ground truth and system track bounding box overlap across all matches and all frames:

$$MOTP = \sum_{t,k} d_{t,k} / \sum_t c_t \quad (2)$$

where $c_t$ is the number of matches and $d_{k,t}$ is the spatial overlap of the $k^{th}$ matched system track in frame $t$. *MOTA* measures the overall accuracy of a system in tracking object configurations together with its ability to preserve object IDs across multiple frames:

$$MOTA = 1 - \sum_t (FN_t + FP_t + IDSW_t) / \sum_t GT_t \quad (3)$$

where, $GT_t$ is the number of unique ground truth objects, $FN_t$ is the number of false negatives, $FP_t$ is the number of false positives and $IDSW_t$ is the number of ID switches.

As an example of its application, Stamatescu et al.[14] applied the CLEAR MOT formalism to evaluate multi-object tracking performance using Neovision2 Tower sequence 001 for all objects and for individual object classes.

## 6. WEBSITE DOWNLOAD

The augmented ground truth data is hosted in the following link: http://www.cls-lab.org/data/neovision2-tower-id. The videos for these ground truth data can be downloaded from the original location of Neovision2 Tower data set, which is: http://ilab.usc.edu/neo2/dataset/tower/training/.

## 7. CONCLUSION

This paper has presented new ground truth data for multi-class multi-object tracking based on the Neovision2 Tower data set. This has been implemented by augmenting the existing ground truth annotations with unique object IDs. A comparison with existing benchmarks highlights the significance of the new ground truths. The paper also provides guidance on how these data sets may be combined with the widely used CLEAR MOT metrics to evaluate multi-class multi-object tracking systems. This new data set will allow the research community to evaluate automated vision systems, which perform both the tasks of multi-object tracking and object recognition, in a holistic manner with a single data set.

## ACKNOWLEDGEMENTS

The authors would like to thank Dr. Carmine Pontecorvo for his suggestions, which helped us improve this paper.